\documentclass[10pt,twocolumn,letterpaper]{article}

\usepackage{tikz}
\usetikzlibrary{arrows,shapes,calc,matrix,fit,backgrounds}
\usepackage{pgfplots}
\usepackage{pgfplotstable}
\pgfplotsset{compat=1.9}

\usepackage{xstring}

\usepgfplotslibrary{external}
\newcommand{\extfig}[2]{\tikzsetnextfilename{fig/extern/#1}{#2}}

\tikzstyle{tight} = [inner sep=0pt,outer sep=0pt]

\newcommand{\leg}[1]{\addlegendentry{#1}}

\tikzset{every mark/.append style={solid}}
\pgfplotsset{
	grid=both, width=\columnwidth, try min ticks=5,
	every axis/.append style={font=\scriptsize},
	every axis plot/.append style={thick,mark=none,mark size=1.2,tension=0.18},
	legend cell align=left, legend style={fill opacity=0.8},
}

\pgfplotsset{
	dash/.style={mark=o,dashed,opacity=0.7},
	dott/.style={mark=o,dotted,opacity=0.7},
}

\usepackage{iccv}
\usepackage{times}
\usepackage{epsfig}
\usepackage{graphicx}
\usepackage{amsmath}
\usepackage{amssymb}
\usepackage{multirow,bbm}
\usepackage{enumitem}
\usepackage[bottom]{footmisc}

\usepackage[pagebackref=true,breaklinks=true,letterpaper=true,colorlinks,bookmarks=false]{hyperref}

\iccvfinalcopy 

\def\iccvPaperID{3537} 

\ificcvfinal\fi
\begin{document}

\title{Smooth Adversarial Examples}

\author{
Hanwei Zhang \ \ \ \
Yannis Avrithis \ \ \ \
Teddy Furon \ \ \ \
Laurent Amsaleg \\[3pt]
{\fontsize{11}{13}\selectfont Univ Rennes, Inria, CNRS, IRISA}
}

\maketitle

\newcommand{\fig}[2][1]{\includegraphics[width=#1\columnwidth]{fig/#2}}


\newcommand{\head}[1]{{\smallskip\noindent\bf #1}}
\newcommand{\alert}[1]{{\color{red}{#1}}}
\newcommand{\eq}[1]{(\ref{eq:#1})\xspace}

\newcommand{\red}[1]{{\color{red}{#1}}}
\newcommand{\blue}[1]{{\color{blue}{#1}}}
\newcommand{\green}[1]{{\color{green}{#1}}}
\newcommand{\gray}[1]{{\color{gray}{#1}}}


\newcommand{\tran}{^\top}
\newcommand{\mtran}{^{-\top}}
\newcommand{\zcol}{\mathbf{0}}
\newcommand{\zrow}{\zcol\tran}

\newcommand{\ind}{\mathbbm{1}}
\newcommand{\expect}{\mathbb{E}}
\newcommand{\nat}{\mathbb{N}}
\newcommand{\zahl}{\mathbb{Z}}
\newcommand{\real}{\mathbb{R}}
\newcommand{\proj}{\mathbb{P}}
\newcommand{\prob}{\mathbf{Pr}}

\newcommand{\mif}{\textrm{if }}
\newcommand{\minimize}{\textrm{minimize }}
\newcommand{\maximize}{\textrm{maximize }}
\newcommand{\st}{\textrm{subject to }}

\newcommand{\id}{\operatorname{id}}
\newcommand{\const}{\operatorname{const}}
\newcommand{\sgn}{\operatorname{sgn}}
\newcommand{\var}{\operatorname{Var}}
\newcommand{\mean}{\operatorname{mean}}
\newcommand{\trace}{\operatorname{tr}}
\newcommand{\diag}{\operatorname{diag}}
\newcommand{\vect}{\operatorname{vec}}
\newcommand{\cov}{\operatorname{cov}}

\newcommand{\softmax}{\operatorname{softmax}}
\newcommand{\clip}{\operatorname{clip}}

\newcommand{\defn}{\mathrel{:=}}
\newcommand{\peq}{\mathrel{+\!=}}
\newcommand{\meq}{\mathrel{-\!=}}

\newcommand{\floor}[1]{\left\lfloor{#1}\right\rfloor}
\newcommand{\ceil}[1]{\left\lceil{#1}\right\rceil}
\newcommand{\inner}[1]{\left\langle{#1}\right\rangle}
\newcommand{\norm}[1]{\left\|{#1}\right\|}
\newcommand{\frob}[1]{\norm{#1}_F}
\newcommand{\card}[1]{\left|{#1}\right|\xspace}
\newcommand{\diff}{\mathrm{d}}
\newcommand{\der}[3][]{\frac{d^{#1}#2}{d#3^{#1}}}
\newcommand{\pder}[3][]{\frac{\partial^{#1}{#2}}{\partial{#3^{#1}}}}
\newcommand{\ipder}[3][]{\partial^{#1}{#2}/\partial{#3^{#1}}}
\newcommand{\dder}[3]{\frac{\partial^2{#1}}{\partial{#2}\partial{#3}}}

\newcommand{\wb}[1]{\overline{#1}}
\newcommand{\wt}[1]{\widetilde{#1}}

\def\xssp{\hspace{1pt}}
\def\ssp{\hspace{3pt}}
\def\msp{\hspace{5pt}}
\def\lsp{\hspace{12pt}}

\newcommand{\cA}{\mathcal{A}}
\newcommand{\cB}{\mathcal{B}}
\newcommand{\cC}{\mathcal{C}}
\newcommand{\cD}{\mathcal{D}}
\newcommand{\cE}{\mathcal{E}}
\newcommand{\cF}{\mathcal{F}}
\newcommand{\cG}{\mathcal{G}}
\newcommand{\cH}{\mathcal{H}}
\newcommand{\cI}{\mathcal{I}}
\newcommand{\cJ}{\mathcal{J}}
\newcommand{\cK}{\mathcal{K}}
\newcommand{\cL}{\mathcal{L}}
\newcommand{\cM}{\mathcal{M}}
\newcommand{\cN}{\mathcal{N}}
\newcommand{\cO}{\mathcal{O}}
\newcommand{\cP}{\mathcal{P}}
\newcommand{\cQ}{\mathcal{Q}}
\newcommand{\cR}{\mathcal{R}}
\newcommand{\cS}{\mathcal{S}}
\newcommand{\cT}{\mathcal{T}}
\newcommand{\cU}{\mathcal{U}}
\newcommand{\cV}{\mathcal{V}}
\newcommand{\cW}{\mathcal{W}}
\newcommand{\cX}{\mathcal{X}}
\newcommand{\cY}{\mathcal{Y}}
\newcommand{\cZ}{\mathcal{Z}}

\newcommand{\vA}{\mathbf{A}}
\newcommand{\vB}{\mathbf{B}}
\newcommand{\vC}{\mathbf{C}}
\newcommand{\vD}{\mathbf{D}}
\newcommand{\vE}{\mathbf{E}}
\newcommand{\vF}{\mathbf{F}}
\newcommand{\vG}{\mathbf{G}}
\newcommand{\vH}{\mathbf{H}}
\newcommand{\vI}{\mathbf{I}}
\newcommand{\vJ}{\mathbf{J}}
\newcommand{\vK}{\mathbf{K}}
\newcommand{\vL}{\mathbf{L}}
\newcommand{\vM}{\mathbf{M}}
\newcommand{\vN}{\mathbf{N}}
\newcommand{\vO}{\mathbf{O}}
\newcommand{\vP}{\mathbf{P}}
\newcommand{\vQ}{\mathbf{Q}}
\newcommand{\vR}{\mathbf{R}}
\newcommand{\vS}{\mathbf{S}}
\newcommand{\vT}{\mathbf{T}}
\newcommand{\vU}{\mathbf{U}}
\newcommand{\vV}{\mathbf{V}}
\newcommand{\vW}{\mathbf{W}}
\newcommand{\vX}{\mathbf{X}}
\newcommand{\vY}{\mathbf{Y}}
\newcommand{\vZ}{\mathbf{Z}}

\newcommand{\va}{\mathbf{a}}
\newcommand{\vb}{\mathbf{b}}
\newcommand{\vc}{\mathbf{c}}
\newcommand{\vd}{\mathbf{d}}
\newcommand{\ve}{\mathbf{e}}
\newcommand{\vf}{\mathbf{f}}
\newcommand{\vg}{\mathbf{g}}
\newcommand{\vh}{\mathbf{h}}
\newcommand{\vi}{\mathbf{i}}
\newcommand{\vj}{\mathbf{j}}
\newcommand{\vk}{\mathbf{k}}
\newcommand{\vl}{\mathbf{l}}
\newcommand{\vm}{\mathbf{m}}
\newcommand{\vn}{\mathbf{n}}
\newcommand{\vo}{\mathbf{o}}
\newcommand{\vp}{\mathbf{p}}
\newcommand{\vq}{\mathbf{q}}
\newcommand{\vr}{\mathbf{r}}
\newcommand{\Vs}{\mathbf{s}}
\newcommand{\vt}{\mathbf{t}}
\newcommand{\vu}{\mathbf{u}}
\newcommand{\vv}{\mathbf{v}}
\newcommand{\vw}{\mathbf{w}}
\newcommand{\vx}{\mathbf{x}}
\newcommand{\vy}{\mathbf{y}}
\newcommand{\vz}{\mathbf{z}}

\newcommand{\vone}{\mathbf{1}}
\newcommand{\vzero}{\mathbf{0}}

\newcommand{\valpha}{{\boldsymbol{\alpha}}}
\newcommand{\vbeta}{{\boldsymbol{\beta}}}
\newcommand{\vgamma}{{\boldsymbol{\gamma}}}
\newcommand{\vdelta}{{\boldsymbol{\delta}}}
\newcommand{\vepsilon}{{\boldsymbol{\epsilon}}}
\newcommand{\vzeta}{{\boldsymbol{\zeta}}}
\newcommand{\veta}{{\boldsymbol{\eta}}}
\newcommand{\vtheta}{{\boldsymbol{\theta}}}
\newcommand{\viota}{{\boldsymbol{\iota}}}
\newcommand{\vkappa}{{\boldsymbol{\kappa}}}
\newcommand{\vlambda}{{\boldsymbol{\lambda}}}
\newcommand{\vmu}{{\boldsymbol{\mu}}}
\newcommand{\vnu}{{\boldsymbol{\nu}}}
\newcommand{\vxi}{{\boldsymbol{\xi}}}
\newcommand{\vomikron}{{\boldsymbol{\omikron}}}
\newcommand{\vpi}{{\boldsymbol{\pi}}}
\newcommand{\vrho}{{\boldsymbol{\rho}}}
\newcommand{\vsigma}{{\boldsymbol{\sigma}}}
\newcommand{\vtau}{{\boldsymbol{\tau}}}
\newcommand{\vupsilon}{{\boldsymbol{\upsilon}}}
\newcommand{\vphi}{{\boldsymbol{\phi}}}
\newcommand{\vchi}{{\boldsymbol{\chi}}}
\newcommand{\vpsi}{{\boldsymbol{\psi}}}
\newcommand{\vomega}{{\boldsymbol{\omega}}}

\newcommand{\rLambda}{\mathrm{\Lambda}}
\newcommand{\rSigma}{\mathrm{\Sigma}}


\newcommand{\our}{\alert{SAE}\xspace}
\newcommand{\func}[1]{\mathsf{#1}}
\newcommand{\magn}{\func{mag}}

\def \Nsuc {N_{\text{suc}}}
\def \Psuc {P_{\text{suc}}}
\def \Xsuc {X_{\text{suc}}}
\def \Dist {\overline{D}}
\def \dist {D}

\def \xa {\vx_a}
\def \xo {\vx_o}

\def \lab {\func{t}}
\def \class {\func{p}}
\def \ff {\func{f}}
\def \fa {\func{a}}
\def \fb {\func{b}}
\def \fc {\func{c}}
\def \fs {\func{s}}
\def \oc {{\func{P}}}

\def \fgsm {FGSM\xspace}
\def \ifgsm {I-FGSM\xspace}
\def \gd {PGD$_2$\xspace}
\def \cw {C\&W\xspace}
\def \scw {sC\&W\xspace}
\def \sgd {sPGD$_2$\xspace}

\def \citeme {\alert{[??]}\xspace}
\def \simple {{C4}\xspace}

\begin{abstract}
This paper investigates the visual quality of the adversarial examples. Recent papers propose to smooth the perturbations to get rid of high frequency artefacts. In this work, smoothing has a different meaning as it perceptually shapes the perturbation according to the visual content of the image to be attacked. The perturbation becomes locally smooth on the flat areas of the input image, but it may be noisy on its textured areas and sharp across its edges.

This operation relies on Laplacian smoothing, well-known in graph signal processing, which we integrate in the attack pipeline. We benchmark several attacks with and without smoothing under a white-box scenario and evaluate their transferability. Despite the additional constraint of smoothness, our attack has the same probability of success at lower distortion.
\end{abstract}

\vspace{-0.5cm}
\section{Introduction}
\label{sec:intro}

\emph{Adversarial examples} where introduced by Szegedy \etal~\cite{szegedy2013intriguing} as \emph{imperceptible} perturbations of a test image that can change a neural network's prediction. This has spawned active research on adversarial attacks and defenses with
competitions among research teams~\cite{1804.00097}. Despite the theoretical and practical progress in understanding the sensitivity of neural networks to their input, assessing the imperceptibility of adversarial attacks remains elusive: user studies show that $L_p$ norms are largely unsuitable, whereas more sophisticated measures are limited too~\cite{ShBR18}.

Machine assessment of perceptual similarity between two images (the input image and its adversarial example) is arguably as difficult as the original classification task, while human assessment of whether one image is adversarial is hard when the $L_p$ norm of the perturbation is small. Of course, when both images are available and the perturbation is isolated, one can always see it. To make the problem interesting, we ask the following question: \emph{given a single image, can the effect of a perturbation be magnified to the extent that it becomes visible and a human may decide whether this example is benign or adversarial?}

\newcommand{\magr}[1]{\extfig{mag-#1}{
\tikz[scale=.215]{
	\node[tight,transform shape](a){\fig[1]{magnify/data/#1}};
	\draw[red,line width=1pt] (-3,-1.5) rectangle (3,1.5);
}}}
\newcommand{\magz}[1]{\extfig{mag-#1}{
\tikz[scale=.6]{
	\clip (-3,-1.5) rectangle (3,1.5);
	\node[tight,transform shape](a){\fig[1]{magnify/data/#1}};
}}}

\begin{figure}
\centering
\begin{tabular}{cc}
	\magr{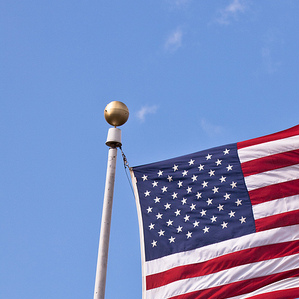} &
	\magz{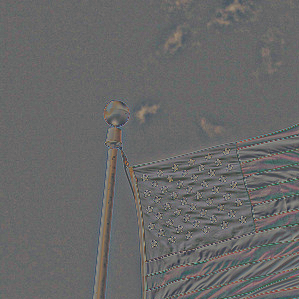} \\
	(a) Original image &
	(b) Original, magnified \\
	\magz{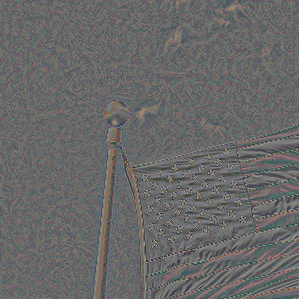} &
	\magz{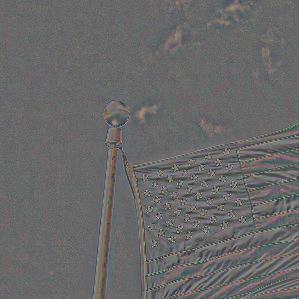} \\
	(c) \cw~\cite{carlini2017towards} &
	(d) \scw\ (this work) \\
	\magz{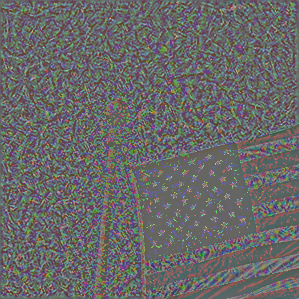} &
	\magz{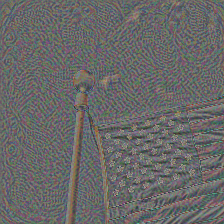} \\
	(e) DeepFool~\cite{moosavi2016deepfool} &
	(f) Universal~\cite{moosavi2017universal} \\
	\magz{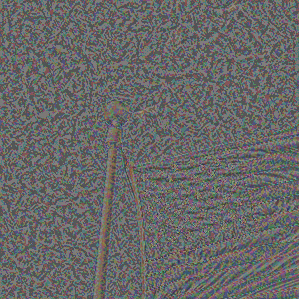} &
	\magz{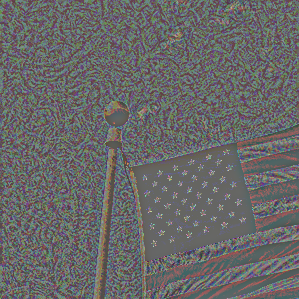} \\
	(g) \fgsm~\cite{goodfellow2014explaining} &
	(h) \ifgsm~\cite{kurakin2016physical}
\end{tabular}
\caption{Given a single input image, our \emph{adversarial magnification} (\cf Appendix~\ref{sec:magnify}) reveals the effect of a potential adversarial perturbation. We show (a) the original image followed by (b) its own magnified version as well as (c)-(h) magnified versions of adversarial examples generated by different attacks. Our \emph{smooth adversarial example} (d) is invisible even when magnified.}
\label{fig:magnify}
\end{figure}

Figure~\ref{fig:magnify} shows that the answer is positive for a range of popular adversarial attacks. In Appendix~\ref{sec:magnify} we propose a simple \emph{adversarial magnification} producing a ``magnified'' version of a given image, without the knowledge of any other reference image. Assuming that natural images are locally smooth, this can reveal not only the existence of an adversarial perturbation but also its pattern. One can recognize, for instance, the pattern of Fig.~4 of~\cite{moosavi2017universal} in our Fig.~\ref{fig:magnify}(f), revealing a universal adversarial perturbation.

Motivated by this example, we argue that popular adversarial attacks have a fundamental limitation in terms of imperceptibility that we attempt to overcome by introducing \emph{smooth adversarial examples}. Our attack assumes local smoothness and generates examples that are consistent with the precise smoothness pattern of the input image. More than just looking ``natural''~\cite{ZDS18} or being smooth~\cite{1807.10590,1809.08758}, our adversarial examples are photorealistic, low-distortion, and virtually invisible even under magnification. This is evident by comparing our magnified example in Fig.~\ref{fig:magnify}(d) to the magnified original in Fig.~\ref{fig:magnify}(b).

Given that our adversarial examples are more constrained, an interesting question is whether they perform well according to metrics like probability of success and $L_p$ distortion. We show that our attack is not only competitive but outperforms Carlini \& Wagner~\cite{carlini2017towards}, from which our own attack differs basically by a smoothness penalty.

\head{Contributions.} As \emph{primary} contributions, we
\begin{enumerate}[noitemsep,topsep=1pt]
	\item investigate the behavior of existing attacks when perturbations become ``\emph{smooth like}'' the input image; and
	\item devise one attack that performs well on standard metrics while satisfying the new constraint.
\end{enumerate}
As \emph{secondary} contributions, we
\begin{enumerate}[noitemsep,topsep=1pt]
	\item[3.] \emph{magnify} perturbations to facilitate qualitative evaluation of their imperceptibility; and
	\item[4.] define a new, more complete/fair \emph{evaluation protocol}.
\end{enumerate}

The remaining text is organized as follows. Section~\ref{sec:related} formulates the problem and introduces a classification of attacks. It describes the C\&W attack and the related work. Section~\ref{sec:background} explains Laplacian smoothing, on which we build our method. Section~\ref{sec:method} presents our \emph{smooth adversarial attacks}, and section~\ref{sec:exp} provides experimental evaluation. Conclusions are drawn section~\ref{sec:discussion}. Our \emph{adversarial magnification} used to generate Fig.~\ref{fig:magnify} is specified in Appendix~\ref{sec:magnify}.

\section{Problem formulation and related work}
\label{sec:related}

Let us denote by $\vx \in \cX \defn [0,1]^{n \times d}$ an image of $n$ pixels and $d$ color channels that has been flattened in a given ordering of the spatial components. A classifier network $\ff$ maps that input image $\vx$ to an output $\vy=\ff(\vx) \in \real^k$ which contains the \emph{logits} of $k$ classes. It is typically followed by $\softmax$ and cross-entropy loss at supervised training or by $\arg\max$ at test time. An input $\vx$ with logits $\vy = \ff(\vx)$ is correctly classified if the \emph{prediction} $\class(\vx)\defn \arg \max_i y_i$ equals the true label of $\vx$.

The attacker mounts a \emph{white-box} attack that is specific to $\ff$, public and known. The attack modifies an original image $\xo \in \cX$ with given true label $t \in \{1,\dots,k\}$ into an adversarial example $\xa \in \cX$, which may be incorrectly classified by the network, that is $\class(\xa) \ne t$, although it looks similar to the original $\xo$. The latter is often expressed by a small $L_2$ \emph{distortion} $\norm{\xa-\xo}$.

\subsection{Families of attacks}
\label{sec:Families}

In a white box setting, attacks typically rely on exploiting the gradient of some loss function. We propose to classify known attacks into three families.

\head{Target Distortion}. This family gathers attacks targeting a distortion $\epsilon$ given as an input parameter. Examples are early attacks like Fast Gradient Sign Method (\fgsm)~\cite{goodfellow2014explaining} and Iterative-\fgsm\ (\ifgsm)~\cite{kurakin2016physical}. Their performance is then measured by the \emph{probability of success} $\Psuc \defn \mathbbm{P}(\class(\xa)\neq t)$ as a function of $\epsilon$.

\head{Target Success}. This family gathers attacks that always succeed in misclassifying $\xa$, at the price of a possible large distortion. DeepFool~\cite{moosavi2016deepfool} is a typical example. Their performance is then measured by the \emph{expected distortion} $\Dist \defn \mathbbm{E}(\|\xa - \xo\|)$.

These two first families are implemented with variations of a gradient descent method. A \emph{classification loss} function is defined on an output logit vector $\vy = \ff(\vx)$ with respect to the original true label $t$, denoted by $\ell(\vy,t)$.

\head{Target Optimality}. The above attacks are not optimal because they a priori do not solve the problem of succeeding under minimal distortion,
\begin{equation}
\min_{\vx\in \cX: \class(\vx)\neq t} \|\vx - \xo\|.
\end{equation}
Szegedy \etal~\cite{szegedy2013intriguing} approximate this constrained minimization problem by a Lagrangian formulation
\begin{equation}
	\min_{\vx\in \cX}  \lambda \norm{\vx-\xo}^2 + \ell(\ff(\vx),t). \label{eq:lbfgs}
\end{equation}
Parameter $\lambda$ controls the trade-off between the distortion and the classification loss.
Szegedy \etal~\cite{szegedy2013intriguing} carry out this optimization by box-constrained L-BFGS.

The attack of Carlini \& Wagner~\cite{carlini2017towards}, denoted \cw in the sequel, pertains to this approach. A change of variable eliminates the box constraint: $\vx\in \cX$ is replaced by $\sigma(\vw)$, where $\vw \in \real^{n \times d}$ and $\sigma$ is the element-wise sigmoid function. A margin is introduced: an \emph{untargeted attack} makes the logit $y_t$ less than any other logit $y_i$ for $i \ne t$ by at least a margin $m\geq0$. Similar to the multi-class SVM loss by Crammer and Singer~\cite{CrSi01} (where $m=1$), the loss function $\ell$ is then defined as
\begin{align}
	\ell(\vy,t) \defn [y_t - \max_{i \ne t} y_i + m]_+,
\label{eq:loss}
\end{align}
where $[\cdot]_+$ denotes the positive part. The \cw attack uses the Adam  optimizer~\cite{KiBa15} to minimize the functional
\begin{equation}
J(\vw,\lambda)\defn\lambda \norm{\sigma(\vw)-\xo}^2 + \ell(\ff(\sigma(\vw)),t).
\label{eq:cw}
\end{equation}
When the margin is reached, loss $\ell(\vy,t)$~\eq{loss} vanishes and the distortion term pulls $\sigma(\vw)$ back towards $\xo$, causing oscillations around the margin. Among all successful iterates, the one with the least distortion is kept; if there is none, the attack fails. The process is repeated for different Lagrangian multiplier $\lambda$ according to line search. This family of attacks is typically more expensive than the two first.

\subsection{Imperceptibility of adversarial perturbations}
Adversarial perturbations are often invisible only because their amplitude is extremely small. Few papers deal with the need of improving the imperceptibility of the adversarial perturbations. The main idea in this direction is to create low or mid-frequency perturbation patterns.

Zhou \etal~\cite{Zhou_2018_ECCV} add a regularization term for the sake of transferability, which removes the high frequencies of the perturbation via low-pass spatial filtering. Heng \etal~\cite{1807.10590} propose a harmonic adversarial attack where perturbations are very smooth gradient-like images. Guo \etal~\cite{1809.08758} design an attack explicitly in the Fourier domain. However, in all cases above, the convolution and the bases of the harmonic functions and of the Fourier transform are independent of the visual content of the input image.

In contrast, the adversarial examples in this work are crafted to be locally compliant with the smoothness of the original image. Our perturbation may be sharp across the edges of $\xo$ but smooth wherever $\xo$ is, \eg on background regions. It is not just smooth but photorealistic, because its smoothness pattern is guided by the input image.

An analogy becomes evident with \emph{digital watermarking}~\cite{Quiring:2018qy}. In this application, the watermark signal pushes the input image into the detection region (the set of images deemed as watermarked by the detector), whereas here the adversarial perturbation drives the image outside its class region. The watermark is invisible thanks to the masking property of the input image~\cite{Cox01-book}. Its textured areas and its contours can hide a lot of watermarking power, but the flat areas can not be modified without producing noticeable artefacts. Perceptually shaping the watermark signal allows a stronger power, which in turn yields more robustness.

Another related problem, with similar solutions mathematically, is photorealistic \emph{style transfer}. Luan \etal~\cite{luan2017deep} transfer style from a reference style image to an input image, while constraining the output to being photorealistic with respect to the input. This work as well as follow-up works~\cite{puy2018flexible,li2018closed} are based on variants of Laplacian smoothing or regularization much like we do.

It is important to highlight that high frequencies can be powerful for deluding a network, as illustrated by the extreme example of the \emph{one pixel attack}~\cite{su2017one}. However this is arguably one of the most visible attacks.

\section{Background on graph Laplacian smoothing}
\label{sec:background}

Popular attacks typically produce noisy patterns that are not found in natural images. They may not be visible at first sight because of their low amplitude, but they are easily detected once magnified (see Fig.~\ref{fig:magnify}). Our objective is to craft an adversarial perturbation that is locally as smooth as the input image, remaining invisible through magnification. This section gives background on \emph{Laplacian smoothing}~\cite{zhou2004learning,KiLL08}, a classical operator in \emph{graph signal processing}~\cite{SaMo13,SNF+13}, which we adapt to images here. Section~\ref{sec:method} uses it generate a smooth perturbation guided by the original input image.

\head{Graph}. Laplacian smoothing builds on a weighted undirected graph whose $n$ vertices correspond to the $n$ pixels of the input image $\xo$. The $i$-th vertex of the graph is associated with feature $\vx_i\in[0,1]^{d}$ that is the $i$-th row of $\xo$, that is, $\xo=[\vx_{1},\ldots,\vx_{n}]\tran$. Matrix $\vp\in\real^{n\times 2}$ denotes the spatial coordinates of the $n$ pixels in the image, and similarly $\vp=[\vp_{1},\ldots,\vp_{n}]\tran$. An edge $(i,j)$ of the graph is associated with weight $w_{ij}\geq 0$, giving rise to an $n \times n$ symmetric adjacency matrix $\vW$, for instance defined as
\begin{align}
	w_{ij} \defn
	\begin{cases}
	\func{k_f}(\vx_i,\vx_j) \func{k_s}(\vp_i,\vp_j), & \text{if } i\neq j\\
	0, & \text{if } i=j
	\end{cases}
\end{align}
for $i,j\in\{1,\dots,n\}$,
where $\func{k_f}$ is a \emph{feature kernel} and $\func{k_s}$ is a \emph{spatial kernel}, both being usually Gaussian or Laplacian. The spatial kernel is typically nonzero only on nearest neighbors, resulting in a sparse matrix $\vW$. We further define the $n \times n$ \emph{degree matrix} $\vD \defn \diag(\vW \vone_n)$ where $\vone_n$ is the all-ones $n$-vector.

\head{Regularization~\cite{zhou2004learning}}. Now, given a new signal $\vz \in \real^{n \times d}$ on this graph, the objective of graph smoothing is to find the output signal $\fs_\alpha(\vz) \defn \arg \min_{\vr\in\real^{n\times d}} \phi_\alpha(\vr,\vz)$ with
\begin{equation}
\phi_\alpha(\vr,\vz) \defn
		\frac{\alpha}{2} \sum_{i,j} w_{ij} \norm{\hat{\vr}_i-\hat{\vr}_j}^2 + (1-\alpha) \norm{\vr-\vz}_F^2
		\label{eq:mrf}
\end{equation}
where $\hat{\vr} \defn \vD^{-1/2} \vr$ and $\norm{\cdot}_F$ is the Frobenius norm. The first summand is the \emph{smoothness term}. It encourages $\hat{\vr}_{i}$ to be close to $\hat{\vr}_{j}$ when $w_{ij}$ is large, \ie when pixels $i$ and $j$ of input $\xo$ are neighbours and similar. This encourages $\vr$ to be smooth wherever $\xo$ is. The second summand is the \emph{fitness term} that encourages $\vr$ to stay close to $\vz$. Parameter $\alpha \in [0,1)$ controls the trade-off between the two.

\head{Filtering}. If we symmetrically normalize matrix $\vW$ as $\cW \defn \vD^{-1/2} \vW \vD^{-1/2}$ and define the $n \times n$ \emph{regularized Laplacian} matrix $\cL_\alpha \defn (\vI_n - \alpha \cW) / (1-\alpha)$, then the expression~\eq{mrf} simplifies to the following quadratic form:
\begin{align}
	\phi_\alpha(\vr,\vz) = (1-\alpha) \trace \left( \vr\tran \cL_\alpha \vr - 2 \vz\tran \vr  + \vz\tran \vz\right).
\end{align}
This reveals, by letting the derivative $\ipder{\phi}{\vr}$ vanish independently per column, that the smoothed signal is simply:
\begin{align}
	\fs_\alpha(\vz) = \cL_\alpha^{-1} \vz.
\label{eq:smooth}
\end{align}
This is possible because matrix $\cL_\alpha$ is positive-definite. Parameter $\alpha$ controls the bandwidth of the smoothing: function $\fs_\alpha$ is the all-pass filter for $\alpha=0$ and becomes a strict `low-pass' filter when $\alpha\to1$~\cite{IAT+18}.

Variants of the model above have been used for instance for interactive image segmentation~\cite{Grad06,KiLL08,Vernaza_2017_CVPR}, transductive semi-supervised classification~\cite{ZhGL03,zhou2004learning}, and ranking on manifolds~\cite{zhou2004ranking,ITA+17}. Input $\vz$ expresses labels known for some input pixels (for segmentation) or samples (for classification), or identifies queries (for ranking), and is null for the remaining vertices. Smoothing then spreads the labels to these vertices according the weights of the graph.

\head{Normalization}. Contrary to applications like interactive segmentation or semi-supervised classification~\cite{zhou2004learning,KiLL08}, $\vz$ does not represent a binary labeling but rather an arbitrary perturbation in this work. Also contrary to such applications, the output is neither normalized nor taken as the maximum over feature dimensions (channels). If $\cL_\alpha^{-1}$ is seen as a spatial filter, we therefore row-wise normalize it to one in order to preserve the dynamic range  of $\vz$:
\begin{align}
	\hat{\fs}_\alpha(\vz) \defn \diag(\fs_\alpha(\vone_n))^{-1} \fs_\alpha(\vz).
\label{eq:norm}
\end{align}
The \emph{normalized smoothing function} $\hat{\fs}_\alpha$ of course depends on $\xo$. We omit this from notation but we say $\hat{\fs}_\alpha$ is \emph{smoothing guided} by $\xo$ and the output is \emph{smooth like} $\xo$.

\section{Integrating smoothness into the attack}
\label{sec:method}

The key idea of the paper is that the smoothness of the perturbation is now consistent with the smoothness of the original input image $\xo$, which is achieved by smoothing operations guided by $\xo$. This section integrates smoothness into attacks targeting distortion (section~\ref{sec:Simple}) and attacks targeting optimality (section~\ref{sec:Families}).

\subsection{Simple attacks}
\label{sec:Simple}
We consider here simple attacks targeting distortion or success based on gradient descent of the loss function. There are many variations which normalize or clip the update according to the norm used for measuring the distortion, a learning rate or a fixed step \etc. These variants are loosely prototyped as the iterative process
\begin{align}
\vg         &= \nabla_{\vx} \ell(\ff(\xa^{(k)}),t), \\
\xa^{(k+1)} &= \fc\left(\xa^{(k)} - \func{n}(\vg)\right),
\end{align}
where $\fc$ is a \emph{clipping} function and $\func{n}$ a \emph{normalization} function according to the variant. Function $\fc$ should at least produce a valid image: $\fc(\vx)\in \cX=[0,1]^{n\times d}$.

\head{Quick and dirty}.
To keep these simple attacks simple, smoothness is loosely integrated after the gradient computation and before the update normalization:
\begin{equation}
\xa^{(k+1)} = \fc\left(\xa^{(k)} - \func{n}(\hat{\fs}_\alpha(\vg))\right).
\label{eq:QuickDirty}
\end{equation}
This approach can be seen as a projected gradient descent on the manifold of perturbations that are smooth like $\xo$.

\subsection{Attack targeting optimality}
\label{sec:Complex}

This section integrates smoothness in the attacks targeting optimality like \cw. Our starting point is the unconstrained problem~\eq{cw}~\cite{carlini2017towards}. However, instead of representing the perturbation signal $\vr \defn \vx-\xo$ implicitly as a function $\sigma(\vw)-\xo$ of another parameter $\vw$, we express the objective explicitly as a function of variable $\vr$, as in the original formulation of~\eq{lbfgs} in~\cite{szegedy2013intriguing}. We make this choice because we need to directly process the perturbation $\vr$ independently of $\xo$. On the other hand, we now need the element-wise clipping function $\fc(\vx) \defn \min([\vx]_+,1)$ to satisfy the constraint $\vx = \xo+\vr \in \cX$~\eq{lbfgs}. Our problem is then
\begin{equation}
	\min_{\vr}  \quad \lambda \norm{\vr}^2 + \ell(\ff(\fc(\xo+\vr)),t),
\label{eq:clip}
\end{equation}
where $\vr$ is unconstrained in $\real^{n \times d}$.

\head{Smoothness penalty}. At this point, optimizing~\eq{clip} results in `independent' updates at each pixel. We would rather like to take the smoothness structure of the input $\xo$ into account and impose a similar structure on $\vr$. Representing the pairwise relations by a graph as discussed in section~\ref{sec:background}, a straightforward choice is to introduce a pairwise loss term
\begin{align}
	\mu \sum_{i,j} w_{ij} \norm{\hat{\vr}_i-\hat{\vr}_j}^2
\label{eq:pair}
\end{align}
into~\eq{clip}, where we recall that $w_{ij}$ are the elements of the adjacency matrix $\vW$ of $\xo$, $\hat{\vr} \defn \vD^{-1/2} \vr$ and $\vD \defn \diag(\vW \vone_n)$. A problem is that the spatial kernel is typically narrow to capture smoothness only locally. Even if parameter $\mu$ is large, it would take a lot of iterations for the information to propagate globally, each iteration needing a forward and backward pass through the network.

\head{Smoothness constraint}. What we advocate instead is to apply a global smoothing process at each iteration: we introduce a \emph{latent variable} $\vz \in \real^{n \times d}$ and seek for a joint solution with respect to $\vr$ and $\vz$ of the following
\begin{equation}
	\min_{\vr,\vz} \quad \mu \phi_\alpha(\vr,\vz) + \lambda \norm{\vr}^2 + \ell(\ff(\fc(\xo+\vr)),t),
\label{eq:joint}
\end{equation}
where $\phi$ is defined by~\eq{mrf}. In words, $\vz$ represents an unconstrained perturbation, while $\vr$ should be close to $\vz$, smooth like $\xo$, small, and such that the perturbed input $\xo+\vr$ satisfies the classification objective. Then, by letting $\mu \to \infty$, the first term becomes a hard constraint imposing a globally smooth solution at each iteration:
\begin{align}
	\min_{\vr,\vz} & \quad \lambda \norm{\vr}^2 + \ell(\ff(\fc(\xo+\vr)),t) \label{eq:global} \\
	\st        & \quad \vr = \hat{\fs}_\alpha(\vz),                     \label{eq:argmin}
\end{align}
where $\hat{\fs}_\alpha$ is defined by~\eq{norm}. During optimization, every iterate of this perturbation $\vr$ is smooth like $\xo$.

\head{Optimization.} With this definition in place, we solve for $\vz$ the following unconstrained problem over $\real^{n\times d}$:
\begin{align}
	\min_{\vz} & \quad \lambda \norm{\hat{\fs}_\alpha(\vz)}^2 + \ell(\ff(\fc(\xo+\hat{\fs}_\alpha(\vz))),t).
\label{eq:sae}
\end{align}
Observe that this problem has the same form as~\eq{clip}, where $\vr$ has been replaced by $\hat{\fs}_\alpha(\vz)$. This implies that we can use the same optimization method as the \cw\ attack. The only difference is that the variable is $\vz$, which we initialize by $\vz=\mathbf{0}_{n\times d}$, and we apply function $\hat{\fs}_\alpha$ at each iteration.

Gradients are easy to compute because our smoothing is a linear operator. We denote the loss on this new variable by $L(\vz) \defn \ell(\ff(\fc(\xo+\hat{\fs}_\alpha(\vz))),t)$. Its gradient is
\begin{equation}
\nabla_{\vz}L(\vz) = \vJ_{\hat{\fs}_\alpha}(\vz)\tran \cdot \nabla_{\vx}\ell(\ff(\fc(\xo+\hat{\fs}_\alpha(\vz))),t),
\end{equation}
where $\vJ_{\hat{\fs}_\alpha}(\vz)$ is the $n\times n$ Jacobian matrix of the smoothing operator at $\vz$.
Since our smoothing operator as defined by~\eq{smooth} and \eq{norm} is linear, $\vJ_{\hat{\fs}_\alpha}(\vz) = \diag(\fs_\alpha(\vone_n))^{-1}\cL_\alpha^{-1}$ is a matrix constant in $\vz$, and multiplication by this matrix is equivalent to smoothing. The same holds for the distortion penalty $\norm{\hat{\fs}_\alpha(\vz)}^2$. This means that in the backward pass, the gradient of the objective~\eq{sae} \wrt $\vz$ is obtained from the gradient \wrt $\vr$ (or $\vx$) by smoothing, much like how $\vr$ is obtained from $\vz$ in the forward pass~\eq{argmin}.

Matrix $\cL_\alpha$ is fixed during optimization, depending only on input $\xo$. For small images like in the MNIST dataset~\cite{LBBH98}, it can be inverted: function $\hat{\fs}_\alpha$ is really a matrix multiplication. For larger images, we use the \emph{conjugate gradient} (CG) method~\cite{NoWr06} to solve the set of linear systems $\cL_\alpha \vr = \vz$ for $\vr$ given $\vz$. Again, this is possible because matrix $\cL_\alpha$ is positive-definite, and indeed it is the most common solution in similar problems~\cite{Grad06,ChKo16,ITA+17}. At each iteration, one computes a product of the form $\vv \mapsto \cL_\alpha \vv$, which is efficient because $\cL_\alpha$ is sparse. In the backward pass, one can either use CG on the gradient, or auto-differentiate through the forward CG iterations. These options have the same complexity. We choose the latter.

\head{Discussion.} The \emph{clipping function} $\fc$ that we use is just the identity over the interval $[0,1]$ but outside this interval its derivative is zero. Carlini \& Wagner~\cite{carlini2017towards} therefore argue that the numerical solver of problem~\eq{clip} suffers from getting stuck in flat spots: when a pixel of the perturbed input $\xo+\vr$ falls outside $[0,1]$, it keeps having zero derivative after that and with no chance of returning to $[0,1]$ even if this is beneficial. This limitation does not apply to our case thanks to the $L_{2}$ distortion penalty in~\eq{clip} and to the updates in its neighborhood: such a value may return to $[0,1]$ thanks to the smoothing operation.

\section{Experiments}
\label{sec:exp}

\begin{table}[t]
\centering
\small
\caption{Success probability $\Psuc$ and average $L_{2}$ distortion $\Dist$.}
\label{tab:2}
\begin{tabular}{|c|c|c|c|c|c|c|}
\hline
\multirow{3}{*}{} & \multicolumn{2}{c|}{MNIST} & \multicolumn{4}{c|}{ImageNet} \\ \cline{2-7}
\multirow{3}{*}{} & \multicolumn{2}{c|}{\simple} & \multicolumn{2}{l|}{InceptionV3} & \multicolumn{2}{l|}{ResNetV2} \\ \cline{2-7}
        & $\Psuc$ & $\Dist$ & $\Psuc$ & $\Dist$ & $\Psuc$ & $\Dist$ \\ \hline \hline
\fgsm   & 0.89    & 5.02    & 0.97    & 5.92    & 0.92    & 8.20    \\ \hline
\ifgsm  & 1.00    & 2.69    & 1.00    & 5.54    & 0.99    & 7.58    \\ \hline
\gd     & 1.00    & 1.71    & 1.00    & 1.80    & 1.00    & 3.63    \\ \hline
\cw     & 1.00    & 2.49    & 0.99    & 4.91    & 0.99    & 9.84    \\ \hline
\hline
\sgd    & 0.97    & 3.36    & 0.96    & 2.10    & 0.93    & 4.80    \\ \hline
\scw    & 1.00    & 1.97    & 0.99    & 3.00    & 0.98    & 5.99    \\ \hline
\end{tabular}
\end{table}

\begin{figure}[t]
\centering
\extfig{pl2-mnist}{
\begin{tikzpicture}
\begin{axis}[
	height=6cm,
	xlabel={$\dist$},
	ylabel={$\Psuc$},
	legend columns=1,
	legend style={at={(1,0)},anchor=south east},
	xmin=0,
]
	\addplot[red]   table{fig/eval/cwMnist.txt};   \leg{\cw}
	\addplot[blue]  table{fig/eval/clipMnist.txt}; \leg{\scw}
	\addplot[violet] table{fig/eval/fgsmMnist.txt}; \leg{\fgsm}
	\addplot[brown] table{fig/eval/bimMnist.txt}; \leg{\ifgsm}
	\addplot[cyan] table{fig/eval/biml2Mnist.txt};  \leg{\gd}
	\addplot[black] table{fig/eval/sbimMnist.txt}; \leg{\sgd}
\end{axis}
\end{tikzpicture}
}
\caption[]{Operating characteristics of the attacks over MNIST. Attacks \gd and \sgd
are tested with target distortion $D\in [1,6]$.}
\label{fig:p-l2-mnist}
\end{figure}
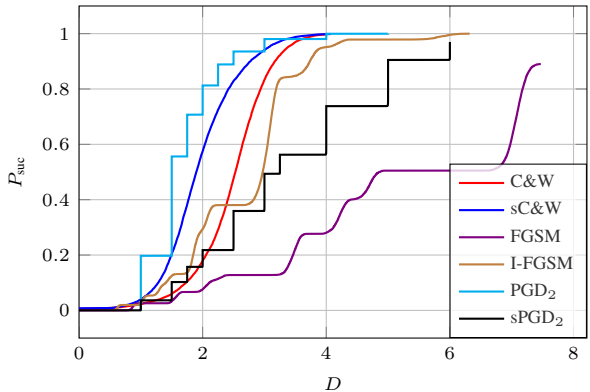

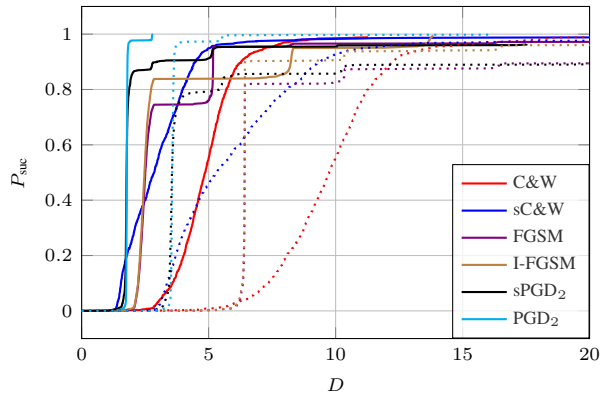
\begin{figure}[t]
\centering
\extfig{pl2-imagenet}{
\begin{tikzpicture}
\begin{axis}[
	height=6cm,
	xlabel={$\dist$},
	ylabel={$\Psuc$},
	legend columns=1,
	legend style={at={(1,0)},anchor=south east},
	xmax=20,
	xmin=0,
]
	\addplot[red]   table{fig/eval/cwImageNet.txt};   \leg{\cw}
	\addplot[blue]  table{fig/eval/clipImageNet.txt}; \leg{\scw}
	\addplot[violet] table{fig/eval/fgsmImageNet.txt}; \leg{\fgsm}
	\addplot[brown] table{fig/eval/bimImageNet.txt};  \leg{\ifgsm}
	\addplot[black] table{fig/eval/sbimImageNet.txt};  \leg{\sgd}
	\addplot[cyan] table{fig/eval/biml2ImageNet.txt};  \leg{\gd}
	\addplot[red,dotted]   table{fig/eval/cwImageNetResV2.txt};   
	\addplot[blue,dotted]  table{fig/eval/clipImageNetResV2.txt}; 
	\addplot[violet,dotted] table{fig/eval/fgsmImageNetResV2.txt}; 
	\addplot[brown,dotted] table{fig/eval/bimImageNetResV2.txt};  
	\addplot[black,dotted]  table{fig/eval/sbimImageNetResV2.txt}; 
	\addplot[cyan,dotted]  table{fig/eval/biml2ImageNetResV2.txt}; 
\end{axis}
\end{tikzpicture}
}
\caption{Operating characteristics over ImageNet attacking InceptionV3 (solid lines) and ResNetV2-50 (dotted lines).}
\label{fig:p-l2-imagenet}
\end{figure}

\begin{figure*}[t]
\centering
\small
\begin{tabular}{cccccc}
	(a) & original image &\cw : $\dist$=3.64 & \scw : $\dist$= 4.59 & \gd :  $\dist$=2.77 & \sgd : $\dist$=5.15
	\\
	&
	\fig[.30]{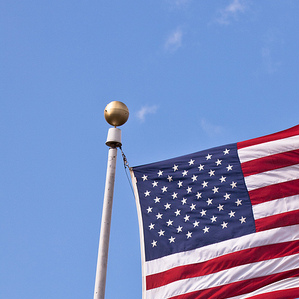} &
	\fig[.30]{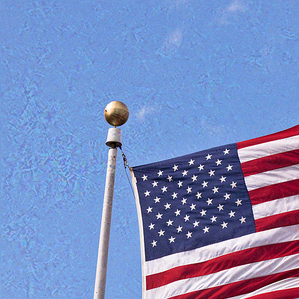} &
	\fig[.30]{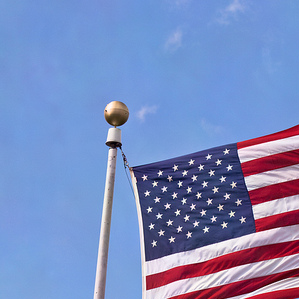} &
	\fig[.30]{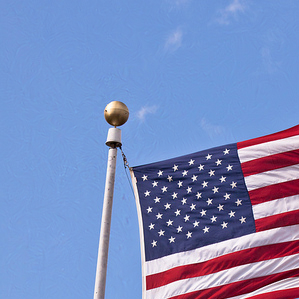} &
	\fig[.30]{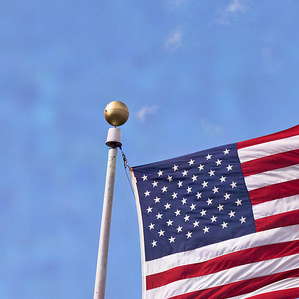} \\
	& &
	\fig[.30]{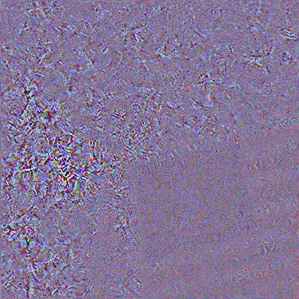} &
	\fig[.30]{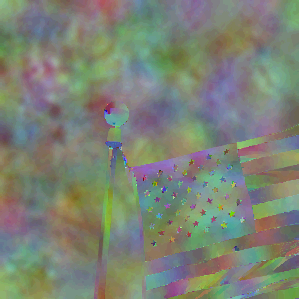} &
	\fig[.30]{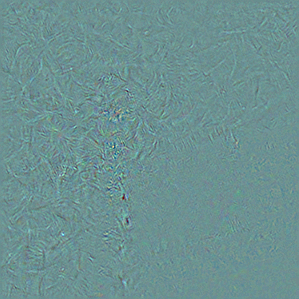} &
	\fig[.30]{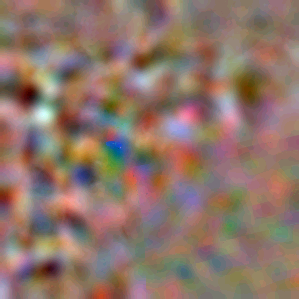} \\
	(b) & original image &\cw : $\dist$=6.55 & \scw : $\dist$= 4.14 & \gd :  $\dist$=2.78 & \sgd : $\dist$=2.82
	\\
	&
	\fig[.30]{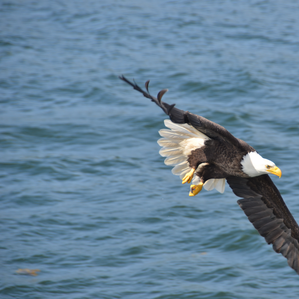} &
	\fig[.30]{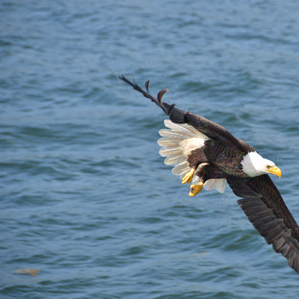} &
	\fig[.30]{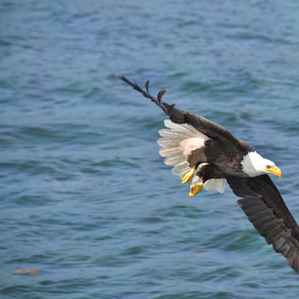} &
	\fig[.30]{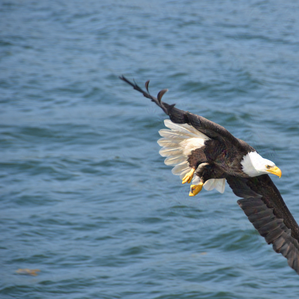} &
	\fig[.30]{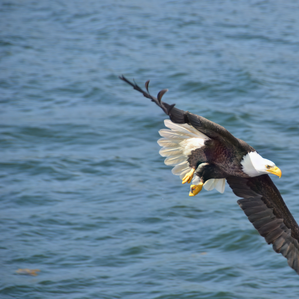} \\
	& &
	\fig[.30]{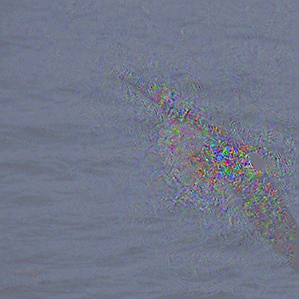} &
	\fig[.30]{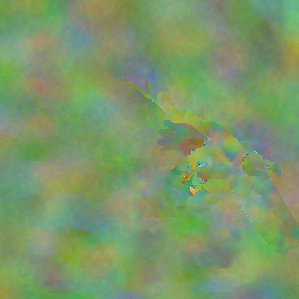} &
	\fig[.30]{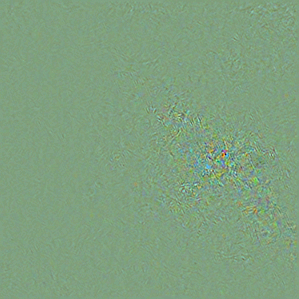} &
	\fig[.30]{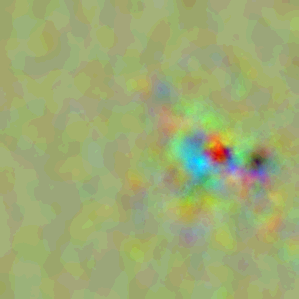} \\
	(c) & original image &\cw : $\dist$=6.55 & \scw : $\dist$= 10.32 & \gd :  $\dist$=2.77 & \sgd : $\dist$=31.76
	\\
	&
	\fig[.30]{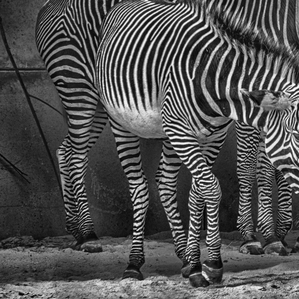} &
	\fig[.30]{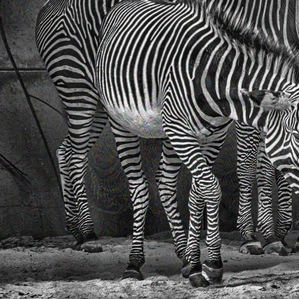} &
	\fig[.30]{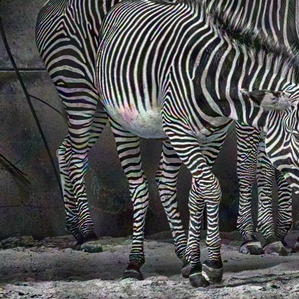} &
	\fig[.30]{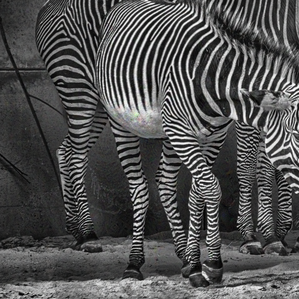} &
	\fig[.30]{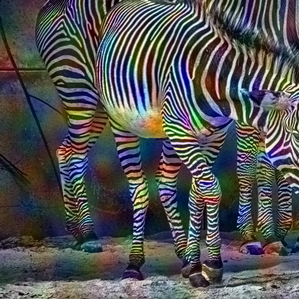} \\
	& &
	\fig[.30]{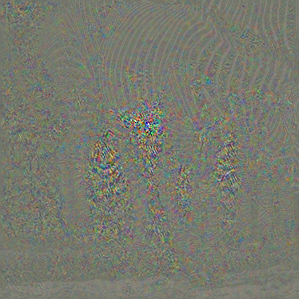} &
	\fig[.30]{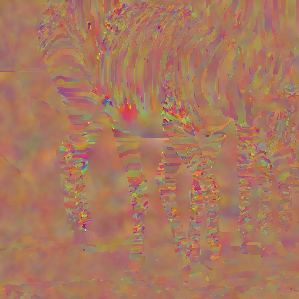} &
	\fig[.30]{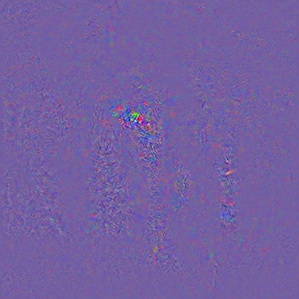} &
	\fig[.30]{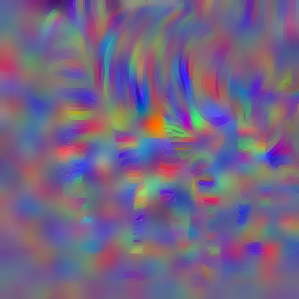} \\
\end{tabular}
\caption{Original image $\xo$ (left), adversarial image $\xa = \xo + \vr$ (above) and scaled perturbation $\vr$ (below; distortion $D=\norm{\vr}$) against InceptionV3 on ImageNet. Scaling maps each perturbation and each color channel independently to $[0,1]$. The perturbation $\vr$ is indeed smooth like $\xo$ for \scw. (a) Despite the higher distortion compared to \cw, the perturbation of \scw is totally invisible, even when magnified (\cf Fig.~\ref{fig:magnify}). (b) One of the failing examples of~\cite{1807.10590} that look unnatural to human vision. (c) One of the examples with the strongest distortion over ImageNet for \scw: $\xo$ is flat along stripes, reducing the dimensionality of the `smooth like $\xo$' manifold.}
\label{fig:ScaledPerturb}
\end{figure*}

Our experiments focus on the \emph{white-box} setting, where the defender first exhibits a network, and then the attacker mounts an attack specific to this network; but we also investigate a \emph{transferability} scenario. All attacks are \emph{untargetted}, as defined by loss function~\eq{loss}.

\subsection{Evaluation protocol}

We evaluate the strength of an attack by two global statistics and by an operating characteristic curve. Given a test image set of $N'$ images, we only consider its subset $X$ of $N$ images that are classified correctly without any attack. The accuracy of the classifier is $N/N'$. Let $\Xsuc$ be the subset of $X$ with $\Nsuc\defn|\Xsuc|$ where the attack succeeds and let $D(\xo) \defn \|\xa-\xo\|$ be the distortion for image $\xo\in\Xsuc$.

The global statistics are the \emph{success probability} $\Psuc$ and \emph{expected distortion} $\Dist$ as defined in section~\ref{sec:related}, estimated by
\begin{equation}
\Psuc = \frac{\Nsuc}{N},\quad\Dist = \frac{1}{\Nsuc}\sum_{\xo \in \Xsuc} D(\xo),
\end{equation}
with the exception that $\Dist$ here is the \emph{conditional average distortion}, where conditioning is on success. Indeed, distortion makes no sense for a failure.

If $D_{\max} = \max_{\xo\in\Xsuc} D(\xo)$ is the maximum distortion, the \emph{operating characteristic} function $\oc: [0,D_{\max}]\to[0,1]$ measures the probability of success as a function of a given upper bound $D$ on distortion. For $D\in[0,D_{\max}]$,
\begin{equation}
\oc(D) \defn \frac{1}{N} |\{\xo \in \Xsuc: D(\xo) \le D\}|.
\label{eq:char}
\end{equation}
This function increases from $\oc(0)=0$ to $\oc(D_{\max}) = \Psuc$.

It is difficult to define a fair comparison of \emph{distortion targeting} attacks to \emph{optimality targeting} attacks. For the first family, we run a given attack several times over the test set with different target distortion $\epsilon$. The attack succeeds on image $\xo \in X$ if it succeeds on any of the runs. For $\xo \in \Xsuc$, the distortion $D(\xo)$ is the minimum distortion over all runs. All statistics are then evaluated as above.

\subsection{Datasets, networks, and attacks}

\head{MNIST~\cite{lecun2010mnist}.} We consider a simple convolutional network with three convolutional layers and one fully connected layer that we denote as \simple, giving accuracy $0.99$. In detail, the first convolutional layer has 64 features, kernel of size $8$ and stride $2$; the second layer has 128 features, kernel of size $6$ and stride $2$; the third has also 128 features, but kernel of size $5$ and stride $1$.

\head{ImageNet.}
We use the dataset of the NIPS 2017 adversarial competition~\cite{kurakin2018adversarial}, comprising 1,000 images from ImageNet~\cite{deng2009imagenet}. We use InceptionV3~\cite{szegedy2016rethinking} and ResNetV2-50~\cite{he2016identity} networks, with accuracy $0.96$ and $0.93$ respectively.

\head{Attacks}. The following six attacks are benchmarked:
\begin{itemize}[noitemsep,topsep=1pt]
\item $L_{\infty}$ distortion: \fgsm~\cite{goodfellow2014explaining} and \ifgsm~\cite{kurakin2016physical}.
\item $L_{2}$ distortion: an $L_2$ version of \ifgsm\ ~\cite{papernot2018cleverhans}, denoted as \gd (projected gradient descent).
\item Optimality: The $L_{2}$ version of \cw~\cite{carlini2017towards}.
\item Smooth: our smooth versions \sgd of \gd (sect.~\ref{sec:Simple}) and \scw\ of \cw\ (sect.~\ref{sec:Complex}).
\end{itemize}

\head{Parameters.}
On MNIST, we use $\epsilon = 0.3$ for \fgsm; $\epsilon = 0.3, \alpha = 0.08$ for \ifgsm; $\epsilon = 5, \alpha = 3$ for \gd; confidence margin $m=1$, learning rate $\eta = 0.1$, and initial constant $c=15$ (the inverse of $\lambda$ in~\eq{cw}) for \cw. For smoothing, we use Laplacian feature kernel, set $\alpha = 0.95$, and pre-compute $\cL_\alpha^{-1}$. On ImageNet, we use $\epsilon = 0.1255$ for \fgsm; $\epsilon = 0.1255, \alpha = 0.08$ for \ifgsm; $\epsilon = 5, \alpha = 3$ for \gd; $m=0$, $\eta = 0.1$, and $c=100$ for \cw. For smoothing, we use Laplacian feature kernel, set $\alpha = 0.997$, and use 50 iterations of CG. These settings are used in sect.~\ref{sec:transf}.

\subsection{White box scenario}

The global statistics $\Psuc,\Dist$ are shown in Table~\ref{tab:2}. Operating characteristics over MNIST and ImageNet are shown in Figures~\ref{fig:p-l2-mnist} and~\ref{fig:p-l2-imagenet} respectively.

We observe that our \scw, with the proper integration via a latent variable~\eqref{eq:sae}, improves a lot the original \cw in terms of distortion, while keeping the probability of success roughly the same. This is surprising. We would expect a price to be paid for a better invisibility as the smoothing is adding an extra constraint on the perturbation. All results clearly show the opposite. On the contrary, the `quick and dirty' integration~\eqref{eq:QuickDirty} dramatically spoils \sgd with big distortion especially on MNIST. This reveals that attacks behave in different ways under the new constraint.

We further observe that \gd outperforms by a vast margin \cw, which is supposed to be close to optimality. This may be due in part to how the Adam optimizer treats $L_2$ norm penalties as studied in~\cite{Loshchilov2017FixingWD}. \cw internally optimizes its parameter $c=1/\lambda$ independently per image, while for \gd we externally try a small set of target distortions $D$ on the entire dataset. This is visible in Fig.~\ref{fig:p-l2-mnist}, where the operating characteristic is piecewise constant. This interesting finding is a result of our new evaluation protocol. Our comparison is fair, given that \cw is more expensive.

As already observed in the literature, ResnetV2 is more robust than InceptionV3: the operating characteristic curves are shifted to the right and increase at a slower rate.

To further understand the different behavior of \cw and \gd under smoothness, Figures~\ref{fig:worst-img-mnist} and~\ref{fig:ScaledPerturb} show MNIST and ImageNet examples respectively, focusing on worst cases. Both \sgd and \scw\ produce smooth perturbations that look more natural. However, smoothing of \sgd is more aggressive especially on MNIST, as these images contain flat black or white areas. The perturbation update $\hat{\fs}_\alpha(\vg)$ is then weakly correlated with gradient $\vg$, which is not efficient to lower the classification loss. In natural images, except for \emph{worst cases} like Fig~\ref{fig:ScaledPerturb}(c), the perturbation of \scw is totally invisible. The reason is the `phantom' of the original that is revealed when the perturbation is isolated.

\begin{figure}[t]
\centering
\begin{tabular}{cccc}
	\includegraphics[width=.2\columnwidth]{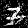} &
	\includegraphics[width=.2\columnwidth]{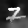} &
	\includegraphics[width=.2\columnwidth]{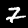} &
	\includegraphics[width=.2\columnwidth]{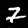} \\
	\textbf{\gd*} &
	\sgd&
	\cw &
	\scw \\
	\green{$\dist$=6.00} &
	\red{$\dist$=6.00} &
	\green{$\dist$=0.36} &
	\green{$\dist$=0.52}\\
	\includegraphics[width=.2\columnwidth]{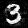} &
	\includegraphics[width=.2\columnwidth]{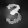} &
	\includegraphics[width=.2\columnwidth]{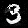} &
	\includegraphics[width=.2\columnwidth]{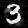} \\
	\gd &
	\textbf{\sgd*} &
	\cw &
	\scw \\

	\green{$\dist$=2.25} &
	\green{$\dist$=6.00} &
	\green{$\dist$=3.33} &
	\green{$\dist$=2.57}\\
	\includegraphics[width=.2\columnwidth]{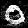} &
	\includegraphics[width=.2\columnwidth]{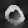} &
	\includegraphics[width=.2\columnwidth]{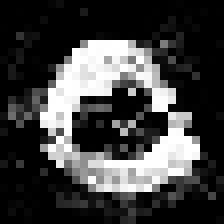} &
	\includegraphics[width=.2\columnwidth]{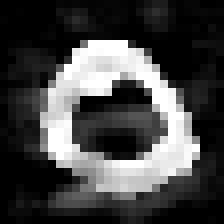} \\

	\gd &
	\sgd &
	\textbf{\cw*} &
	\scw \\

	\green{$\dist$=4.00} &
	\red{$\dist$=6.00} &
	\green{$\dist$=4.22} &
	\green{$\dist$=3.31}\\
	\includegraphics[width=.2\columnwidth]{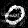} &
	\includegraphics[width=.2\columnwidth]{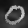} &
	\includegraphics[width=.2\columnwidth]{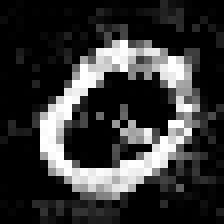} &
	\includegraphics[width=.2\columnwidth]{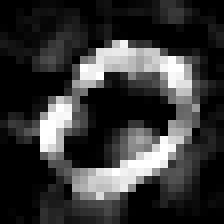} \\
	\gd &
	\sgd &
	\cw &
	\textbf{\scw*}\\
	\green{$\dist$=4.00} &
	\red{$\dist$=6.00} &
	\green{$\dist$=4.15} &
	\green{$\dist$=4.85}\\
\end{tabular}
\caption{For a given attack (denoted by * and bold typeface), the adversarial image with the strongest distortion $\dist$ over MNIST. In green, the attack succeeds; in red, it fails.}
\label{fig:worst-img-mnist}
\end{figure}

\head{Adversarial training}.
The defender now uses adversarial training~\cite{goodfellow2014explaining} to gain robustness against attacks. Yet, the white-box scenario still holds: this network is public. The training set comprises images attacked with ``step l.l'' model~\cite{kurakin2016physical}\footnote{Model taken from: \url{https://github.com/tensorflow/models/tree/master/research/adv_imagenet_models}}. The accuracy of \simple over MNIST (resp. InceptionV3 over ImageNet) is now $0.99$ (resp. $0.94$).

Table~\ref{tab:Adv} shows interesting results.
As expected, \fgsm is defeated in all cases, while average distortion of all attacks is increased in general. What is unexpected is that on MNIST, \scw remains successful while the probability of \cw drops. On ImageNet on the other hand, it is the probability of the smooth versions \sgd and \scw that drops. \ifgsm is also defeated in this case, in the sense that average distortion increases too much.

\begin{table}[t]
\centering
\small
\caption{Success probability and average $L_{2}$ distortion $\Dist$ when attacking networks adversarially trained against \fgsm.}
\label{tab:Adv}
\begin{tabular}{|c|c|c|c|c|}
\hline
& \multicolumn{2}{c|}{MNIST - \simple} & \multicolumn{2}{c|}{ImageNet - InceptionV3} \\ \cline{2-5}
                 & $\Psuc$        & $\Dist$ & $\Psuc$        & $\Dist$                  \\ \hline \hline
\fgsm         &   0.15           & 4.53     &  0.06          & 6.40                \\ \hline
\ifgsm        &   1.00           & 3.48     &  0.97          & 29.94               \\ \hline
\gd           &   1.00           & 2.52     &  1.00          & 3.89             \\ \hline
\cw           &   0.93           & 3.03     &  0.95          & 6.43               \\ \hline
\hline
\sgd          &   0.99           & 2.94     &  0.69          & 7.86            \\ \hline
\scw          &   0.99           & 2.39     &  0.75          & 6.22            \\ \hline
\end{tabular}
\end{table}

\begin{table}[t]
\caption{Success probability and average $L_{2}$ distortion $\Dist$ of attacks on variants of InceptionV3 under transferability.}
\centering
\small
\begin{tabular}{|c|c|c|c|c|}
\hline
              & \multicolumn{2}{l|}{Bilateral filter} & \multicolumn{2}{l|}{Adv. training} \\  \cline{2-5}
              & $\Psuc$        & $\Dist$      & $\Psuc$    & $\Dist$   \\ \hline \hline
\fgsm         & 0.77    & 5.13      &    0.04          &   10.20     \\ \hline
\ifgsm        & 0.82    & 5.12   &     0.02         &     10.10   \\ \hline
\gd           & 1.00    & 5.14        &     0.12           &   10.26\\ \hline
\cw           & 0.82    & 4.75      & 0.02        & 10.21        \\ \hline
\hline
\sgd          & 0.95    & 5.13   &    0.01      &    10.17   \\ \hline
\scw          & 0.68    & 2.91       & 0.01        & 4.63     \\ \hline
\end{tabular}
\label{tab:Transfer}
\end{table}

\subsection{Transferability}
\label{sec:transf}

This section investigates the transferability of the attacks under the
following scenario: the attacker has now a partial knowledge about the network.
For instance, he/she knows that the defender chose a variant of InceptionV3, but this variant is not public so he/she attacks InceptionV3 instead. Also, this time he/she is not allowed to test different distortion targets. The results are shown in Table~\ref{tab:Transfer}.

The first variant uses a bilateral filter (with standard deviation 0.5 and 0.2 in the domain and range kernel respectively; \cf appendix~\ref{sec:magnify}) before feeding the network. This does not really prevent the attacks. \gd remains a very powerful attack if the distortion is large enough. Smoothing does not improve the statistics but the perturbations are less visible.
The second variant uses the adversarially trained InceptionV3, which is, on the contrary, a very efficient counter-measure under this scenario.

\section{Conclusion}
\label{sec:discussion}

Smoothing helps mask the adversarial perturbation, when it is `like' the input image. It allows the attacker to delude more robust networks thanks to larger distortions while still being invisible. However, its impact on transferability is mitigated. While clearly not every attack is improved by such smoothing, which was not our objective, it is impressive how \scw improves upon \cw in terms of distortion and imperceptibility at the same time.

The question raised in the introduction is still open: Fig.~\ref{fig:magnify} shows that a human does not make the difference between the input image and its adversarial example even with magnification. This does not prove that an algorithm will not detect some statistical evidence.

\appendix
\section{Adversarial magnification}
\label{sec:magnify}

Given a single-channel image $\vx: \Omega \to \real$ as input, its \emph{adversarial magnification} $\magn(\vx): \Omega \to \real$ is defined as the following \emph{local normalization} operation
\begin{equation}
	\magn(\vx) \defn
		\frac{\vx - \mu_{\vx}(\vx)}{\beta \sigma_\vx(\vx) + (1-\beta) \sigma_\Omega(\vx)},
\label{eq:mag}
\end{equation}
where $\mu_{\vx}(\vx)$ and $\sigma_{\vx}(\vx)$ are the local mean and standard deviation of $\vx$ respectively, and  $\sigma_\Omega(\vx)\in\real^{+}$ is the global standard deviation of $\vx$ over $\Omega$. Parameter $\beta\in[0,1]$ determines how much local variation is magnified in $\vx$.

In our implementation, $\mu_{\vx}(\vx) = \fb(\vx)$, the \emph{bilateral filtering} of $\vx$~\cite{tomasi1998bilateral}. It applies a local kernel at each point $p \in \Omega$ that is the product of a domain and a range Gaussian kernel. The \emph{domain kernel} measures the geometric proximity of every point $q \in \Omega$ to $p$ as a function of $\norm{p-q}$ and the \emph{range kernel} measures the photometric similarity of every point $q \in \Omega$ to $p$ as a function $|\vx(p)-\vx(q)|$. On the other hand, $\sigma_{\vx}(\vx) = \fb_{\vx}((\vx-\mu_{\vx}(\vx))^2)^{-1/2}$, where $\fb_{\vx}$ is a \emph{guided} version of the bilateral filter, where it is the reference image $\vx$ rather than $(\vx-\mu_{\vx}(\vx))^2$ that is used in the range kernel.

When $\vx: \Omega \to \real^d$ is a $d$-channel image, we apply all the filters independently per channel, but photometric similarity is just one scalar per point as a function of the Euclidean distance $\norm{\vx(p)-\vx(q)}$ measured over all $d$ channels.

In Fig.~\ref{fig:magnify}, $\beta = 0.8$. The standard deviation of both the domain and range Gaussian kernels is $5$.

{\small
\bibliographystyle{ieee}
\bibliography{paper}
}

\end{document}